\documentclass{article}
\usepackage[preprint,nonatbib]{nips_2018}

\usepackage{times}
\usepackage{xcolor}
\usepackage{soul}
\usepackage[utf8]{inputenc} 
\usepackage[T1]{fontenc}    
\usepackage[bf, small]{caption}
\usepackage{amsmath}
\usepackage{algorithm}
\usepackage{algpseudocode}
\usepackage{graphicx}
\usepackage{hyperref}       
\usepackage{url}            
\usepackage{booktabs}       

\newcommand{\BIGO}{\mathcal{O}}
\newcommand{\bfx}{\mathbf{x}}
\newcommand{\bfy}{\mathbf{y}}
\newcommand{\bfu}{\mathbf{u}}
\newcommand{\KXX}{K_{XX}}
\newcommand{\TKXX}{\tilde{K}_{XX}}
\newcommand{\KXU}{K_{XU}}
\newcommand{\TKXU}{\tilde{K}_{XU}}
\newcommand{\KUU}{K_{UU}}
\newcommand{\bfxt}{\mathbf{x}_\star}

\newcommand*\samethanks[1][\value{footnote}]{\footnotemark[#1]}

\title{Kernel Distillation for Fast Gaussian Processes Prediction}

\author{
Congzheng Song\thanks{Equal contribution}
\\ 
Cornell Tech\\
cs2296@cornell.edu
\And 
Yiming Sun\samethanks \\
Cornell University\\
ys784@cornell.edu
}

\begin{document}

\maketitle

\begin{abstract}
Gaussian processes (GPs) are flexible models that can capture complex structure in large-scale dataset due to their non-parametric nature. However, the usage of GPs in real-world application is limited due to their high computational cost at inference time.  In this paper, we introduce a new framework, \textit{kernel distillation}, to approximate a fully trained teacher GP model with kernel matrix of size $n\times n$ for $n$ training points. We combine inducing points method with sparse low-rank approximation in the distillation procedure. The distilled student GP model only costs $\BIGO(m^2)$ storage for $m$ inducing points where $m \ll n$ and improves the inference time complexity. We demonstrate empirically that kernel distillation provides better trade-off between the prediction time and the test performance compared to the alternatives.  
\end{abstract}

\section{Introduction}

Gaussian Processes (GPs)~\cite{rasmussen2004gaussian} are powerful tools for regression and classification problems as these models are able to learn complex representation of data through expressive covariance kernels. However, the application of GPs in real-world is limited due to their poor scalability during inference time. For a training data of size $n$, GPs requires $\BIGO(n^2)$ computation and storage for inference a single test point. Previous way for scaling GP inference is either through inducing points methods~\cite{seeger2003fast,titsias2009variational,lawrence2003fast} or structure exploitation~\cite{saatcci2012scalable,wilson2014fast}. More recently, the structured kernel interpolation (SKI) framework and KISS-GP~\cite{wilson2015kernel} further improve the scalability of GPs by unifying inducing points methods and structure exploitation. These methods can suffer from degradation of test performance or require input data to have special grid structure. 

All the previous solutions for scaling GPs focus on training GPs from scratch. In this paper, we focus on a different setting where we have enough resource for training exact GPs and want to apply the trained model for inference on resource-limited devices such as mobile phone or robotics~\cite{deisenroth2015gaussian}. We wish to investigate the possibility of compressing a large trained exact GP model to a smaller and faster approximate GP model while preserve the predictive power of the exact model. This paper proposes \textit{kernel distillation}, a general framework to approximate a trained GP model. Kernel distillation extends inducing point methods with insights from SKI framework and utilizes the knowledges from a trained model. 

In particular, we approximate the exact kernel matrix with a sparse and low-rank structured matrix. We formulate kernel distillation as a constrained $F$-norm minimization problem, leading to more accurate kernel approximation compared to previous approximation approaches. Our method is a general purpose kernel approximation method and does not require kernel function to be separable or stationary and input data to have any special structure.  We evaluate our approach on various real-world datasets, and the empirical results evidence that kernel distillation can better preserving the predictive power of a fully trained GP model and improving the speed simultaneously compared to other alternatives.

\section{Kernel Distillation}
\paragraph{Background.}
We focus on GP regression problem. Denote the dataset as $D$ which consists of input feature vectors $X = \{\bfx_1, \dots, \bfx_n\}$ and real-value targets $\mathbf{y} = \{y_1, \dots, y_n\}$. GP models a distribution over functions $f(\bfx)\sim \mathcal{GP}(\mu, k_\gamma)$, where any set of function values forms a joint Gaussian distribution characterized by mean function $\mu(\cdot)$ and kernel function $k_\gamma(\cdot, \cdot)$ where $\gamma$ is the set of  hyper-parameters to be trained. Based on Gaussian Identity, we can arrive at posterior predictive distribution for inference~\cite{rasmussen2006gaussian}:
\begin{align*}
	\mathbf{f_{\star}} \vert X, X_{\star}, \mathbf{y} \sim \mathcal{N}(& K_{X_{\star}X}(\KXX+ \sigma^2 I)^{-1}\bfy, \\ & K_{X_{\star}X_{\star}} - K_{X_{\star}X}(\KXX+\sigma^2 I)^{-1} K_{XX_{\star}}).
\end{align*}
The matrix $K_{X_{\star}X} = k_\gamma(X_\star, X)$ is the covariance measured between $X_\star$ and $X$.  The prediction for mean and variance cost $\BIGO(n)$ in time and $\BIGO(n^2)$ in storage per test point. 

The computational and storage bottleneck is the exact kernel matrix $\KXX$. KISS-GP~\cite{wilson2015kernel} is a inducing point method for approximating the kernel matrix and thus scaling training of GPs. Given a set of $m$ inducing points $U=\{\bfu_1, \dots, \bfu_m\}$, KISS-GP approximates the kernel matrix $\KXX\approx W_X\KUU W^\top_{X}$ where $X$ is locally interpolated with $U$ and $W_X$ is the interpolation weights.

\paragraph{Formulation.} 
The goal of kernel distillation is to compress a fully trained GP model to an approximate GP model to be used for inference on a resource-limited devices. We assume that we have access to a trained exact GP with full kernel matrix $\KXX$ and all the training data $\{X, \bfy\}$ during distillation. Algorithm 1 in Appendix A outlines our distillation procedure.

We propose to use a student kernel matrix with a sparse and low-rank structure, $\TKXX = W\KUU W^\top$ to approximate a fully trained kernel matrix $\KXX$. $W$ is a sparse matrix and $\KUU$ is the covariance evaluated at a set of inducing points $U$. Similar to KISS-GP~\cite{wilson2015kernel}, we approximate $\KXU$ with $\TKXU=W\KUU$. In KISS-GP, $W$ is calculated using cubic interpolation on grid-structured inducing points. The number of inducing points grows exponentially as the dimension of input data grows, limiting KISS-GP applicable to low-dimensional data. Instead of enforcing inducing points $U$ to be on grid, we choose $m$ centroids from the results of K-means clustering $X$ as the inducing points $U$. In addition, we store $U$ in KD-tree $\mathcal{T}$ for fast nearest neighbor search which will be used in later optimization.

In kernel distillation, we find optimal $W$ through a constrained optimization problem. We constrain each row of $W$ to have at most $b$ non-zero entries. We set the objective function to be the $F$-norm error between teacher kernel and student kernel: 
\begin{align*}
\min_{W} \quad &||\KXX - W\KUU W^\top||_F \\
\text{subject to} \quad &||W_{i}||_0 \le b \quad \forall i
\end{align*}
where $||W_{i}||_0$ denotes the number of non-zero entries at row $i$ of $W$.

\paragraph{Initializing $W$.} 
The initial values of $W$ are crucial for the later optimization. We initialize $W$ with optimal solution to $||\KXU - W\KUU||_F$ with the sparsity constraint. More specifically, for each $\mathbf{x_i}$ in $X$, we find its $b$ nearest points in $U$ by querying $\mathcal{T}$. We denote the indices of these $b$ neighbors as $J$. We then initialize each row $W_i$ of $W$ by solving the following linear least square problem:
\begin{align*}
	 \text{min}_{W_i (J)} \quad || W_i(J)\KUU(J)- (\KXU)_i ||_2 
\end{align*}
where $W_i (J)$ denotes the entries in row $W_i$ indexed by $J$  and $\KUU(J)$ denotes the rows of $\KUU$ indexed by $J$. The entries in $W_i$ with index not in $J$ are set to zero.

\paragraph{Optimizing $W$.} After $W$ is initialized, we solve the $F$-norm minimization problem using standard gradient descent. To satisfy the sparsity constraint, in each iteration, we project each row of the gradient $\nabla W$ to $b$-sparse space according the indices $J$, and then update $W$ accordingly.

\paragraph{Fast prediction.}
One direct application of kernel distillation is for fast prediction with approximated kernel matrix. Given a test point $\bfxt$, we follow similar approximation scheme in the distillation at test time where we try to approximate $K_{\bfxt X}$:
\begin{align*}
K_{\bfxt X} \approx \tilde{K}_{\bfxt X} = W_\star \tilde{K}_{UX} = W_\star \KUU W^\top
\end{align*}
where $W_\star$ is forced to be sparse for efficiency. Then the mean and variance prediction can be approximated by:
\begin{align*}
E[f_{\star}]  & \approx  \tilde{K}_{\bfxt X} (\tilde{K}_{XX} + \sigma^2 I)^{-1}\mathbf{y} \approx W_{\star} \KUU W^\top (\tilde{K}_{XX} + \sigma^2 I)^{-1}\mathbf{y} 
\\
& =  W_{\star} \tilde{\pmb{\alpha}}
\\ 
Var[f_{\star}] & \approx  K_{\bfxt\bfxt} - \tilde{K}_{\bfxt X}[\tilde{K}_{XX} + \sigma^2 I]^{-1}\tilde{K}_{X\bfxt} \\
& \approx  K_{\bfxt\bfxt} - W_{\star}\KUU W^\top[\tilde{K}_{XX} + \sigma^2 I]^{-1}W\KUU W_{\star}^\top \\
&=K_{\bfxt\bfxt} -  W_{\star}V  W_{\star}^\top
\end{align*}
where both $\tilde{\pmb{\alpha}} = \KUU W^\top(\tilde{K}_{XX} + \sigma^2 I)^{-1}\mathbf{y}$ and $
V = \KUU W^\top[\tilde{K}_{XX} + \sigma^2 I]^{-1}W\KUU 
$ can be precomputed during distillation.  

To compute $W_\star$ efficiently, we start by finding $b$ nearest neighbors of $\bfxt$ in $U$ (indexed by $J_\star$) and set the entries in $W_\star$ whose indices are not in $J_{\star}$ to 0. For entries with indices in $J_{\star}$, we solve the following least square problem to get the optimal values for $W_\star(J_{\star})$:
\begin{align*}
\min_{W_\star(J_{\star})}\quad || W_\star(J_{\star})\KUU(J_{\star}) - K_{\bfxt U}(J_{\star})||_2. 
\end{align*}
It takes $\BIGO(b\log m)$ to query the nearest neighbors, $\BIGO(b^3)$ to get $W_\star$ and $\BIGO(b)$ and $\BIGO(b^2)$ for mean and variance prediction respectively. The prediction time complexity is $\BIGO(b\log m+b^3)$ in total. As for storage complexity, we need to store precomputed vector for mean prediction and diagonal of matrix for variance prediction which cost $\BIGO(m^2)$. Table \ref{table:time storage complexity} provides comparison of time and storage complexity for different GP approximation approaches. 

\begin{table*}[t]
\centering 
\caption{Time and storage complexity for prediction for FITC, KISS and Distillation. $n$ is the number of training data, $m$ is the number of inducing points, $d$ is the dimension of input data and $b$ is the sparsity constraint in kernel distillation.}
\begin{tabular}{l|r|r|r}
\hline
Methods & Mean Prediction & Variance Prediction & Storage\\
\hline \hline 
FITC~\cite{snelson2005sparse} & $\BIGO(m)$ & $\BIGO(m^2)$& $\BIGO(nm)$\\
KISS-GP~\cite{wilson2015kernel} & $\BIGO(1)$ & $\BIGO(1)$  & $\BIGO(n+4^d)$ \\
Kernel distillation (this work) & $\BIGO(b\log m +b^3)$ & $\BIGO(b\log m +b^3)$ & $\BIGO(m^2)$ \\
\hline
\end{tabular}
\label{table:time storage complexity}
\end{table*}


\section{Experiments}

We evaluate kernel distillation on the ability to approximate the exact kernel, the predictive power and the speed at inference time. In particular, we compare our approach to FITC and KISS-GP as they are the most popular approaches and are closely related to kernel distillation.
The simulation experiments for reconstructing kernel and comparing predictive power are demonstrated in Appendix B. 

\paragraph{Empirical Study.}

We evaluate the performance of kernel distillation on several benchmark regression data sets. A summary of the datasets is given in Table~\ref{tab:smse}. The detailed setup of experiments is in Appendix C.

We start by evaluating how well kernel distillation can preserve the predictive performance of the teacher kernel. The metrics we use for evaluation is the standardized mean square error (SMSE) defined as $\frac{1}{n}\sum^n_{i=1}(y_i - \hat{y}_i)^2 / Var(\bfy)$
for true label $y_i$ and model prediction $\hat{y}_i$. Table~\ref{tab:smse} summarizes the results.  We can see that exact GPs achieve lowest errors on all of the datasets. FITC gets second lowest error on almost all datasets except for Boston Housing. Errors with kernel distillation are very close to FITC while KISS-GP has the largest errors on every dataset. The poor performance of KISS-GP might be resulted from the loss of information through the projection of input data to low dimension.

\begin{table*}[t]
\begin{center}
 \caption{SMSE Results Comparison. $d$ is the dimension of the input data.  Number of inducing points (on 2D grid) for KISS-GP are 4,900, 10K, 90K, 250K, and number of inducing points for FITC and kernel distillation are 70, 200, 1K, 1K for the for datasets respectively. The sparsity $b$ is set to 20 for Boston Housing and 30 for all other datasets.}\begin{tabular}{  l | r | r | r | r | r | r | r }
  \hline
 Dataset 	& $d$ & \# train & \# test  & Exact & FITC & KISS-GP & Distill \\ \hline \hline 
    Boston Housing  & 13 & 455 & 51 & 0.076 & 0.103 & 0.095 & 0.091\\ 
    Abalone & 8 & 3,133 & 1,044 & 0.434 & 0.438 & 0.446  & 0.439 \\ 
    PUMADYM32N & 32 & 7,168 & 1,024 & 0.044 & 0.044 & 1.001 & 0.069\\
    KIN40K & 8 & 10,000 & 30,000 & 0.013 & 0.030 & 0.386 & 0.173\\ \hline
  \end{tabular}
 \label{tab:smse}
 \end{center}
\end{table*}

\begin{figure*}
\centering
\begin{tabular}{cccc}
\includegraphics[width=0.23\textwidth]{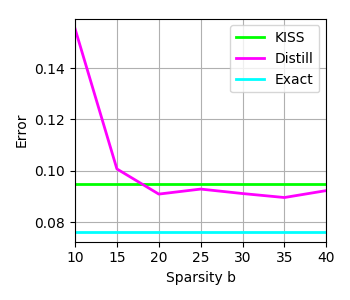} &
\includegraphics[width=0.23\textwidth]{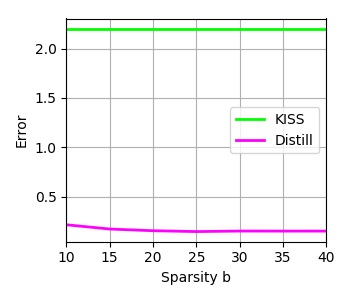} &
\includegraphics[width=0.23\textwidth]{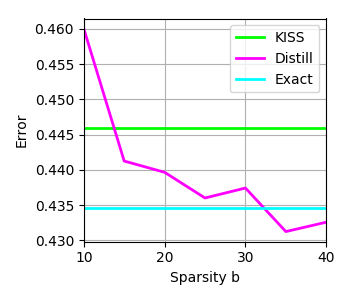} &
\includegraphics[width=0.23\textwidth]{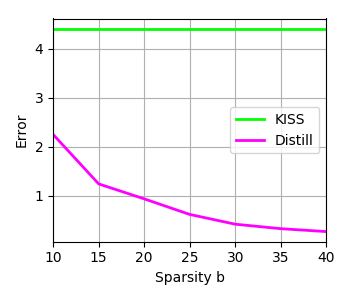} 
\\
(a) Boston Mean & (b) Boston Variance & (c) Abalone Mean & (d) Abalone Variance
\end{tabular}
\caption{Test error and variance comparison on Boston Housing (a-b) and Abalone (c-d) under different choices of sparsity constraint $b$ on $W$. For variance prediction comparison, we calculate the root square mean error between variance of exact GPs and approximate GPs (KISS-GP and kernel distillation).}
\label{fig:sparsity}
\end{figure*}

\begin{table*}
\centering
\caption{Average prediction time in seconds for 1K test data points}
\begin{tabular}{  l | r | r | r }
  \hline
 Dataset & FITC & KISS-GP & Distill \\ \hline \hline 
Boston Housing & 0.0081 & 0.00061 & 0.0017\\
Abalone & 0.0631 & 0.00018 & 0.0020 \\
PUMADYM32N & 1.3414 & 0.0011 & 0.0035 \\
KIN40K & 1.7606 & 0.0029 & 0.0034 \\
\hline
\end{tabular}
\label{tab: average prediction time}
\end{table*}

We further study the effects of sparsity $b$ on predictive performance. We choose $b$ to be range from [5, 10, $\dots$, 40] and compare the test error and variance prediction for KISS-GP and kernel distillation on Boston Housing and Abalone datasets. The results are shown in Figure~\ref{fig:sparsity}. As expected, the error for kernel distillation decreases as the sparsity increases and we only need $b$ to be 15 or 20 to outperform KISS-GP. As for variance prediction, we plot the error between outputs from exact GPs and approximate GPs. We can see that kernel distillation always provides more reliable variance output than KISS-GP on every level of sparsity. 

Finally, we evaluate the speed of prediction with kernel distillation. Again, we compare the speed with FITC and KISS-GP. The setup for the approximate models is the same as the predictive performance comparison experiment. For each dataset, we run the test prediction on 1000 points and report the average prediction time in seconds. Table \ref{tab: average prediction time} summarizes the results on speed. It shows that both KISS-GP and kernel distillation are much faster in prediction time compared to FITC for all datasets. Though kernel distillation is slightly slower than KISS-GP, considering the improvement in accuracy and more reliable uncertainty measurements, the cost in prediction time is acceptable. Also, though KISS-GP claims to have constant prediction time complexity in theory~\cite{wilson2015thoughts}, the actual implementation still is data-dependent and the speed varies on different datasets. In general, kernel distillation provides a better trade-off between predictive power and scalability than its alternatives.


\paragraph{Conclusion.}
We proposed a general framework, kernel distillation, for compressing a trained exact GPs kernel into a student kernel with low-rank and sparse structure. Our framework does not assume any special structure on input data or kernel function, and thus can be applied "out-of-box" on any datasets. Kernel distillation framework formulates the approximation as a constrained $F$-norm minimization between exact teacher kernel and approximate student kernel.

The distilled kernel matrix reduces the storage cost to $\BIGO(m^2)$ compared to $\BIGO(mn)$ for other inducing point methods. Moreover, we show one application of kernel distillation is for fast and accurate GP prediction. Kernel distillation can produce more accurate results than KISS-GP and the prediction time is much faster than FITC. Overall, our method provide a better balance between speed and predictive performance than other approximate GP approaches.

\newpage
\clearpage
\bibliographystyle{unsrt}
\bibliography{ijcai18}

\begin{thebibliography}{10}

\bibitem{rasmussen2004gaussian}
Carl~Edward Rasmussen.
\newblock Gaussian processes in machine learning.
\newblock In {\em Advanced lectures on machine learning}, pages 63--71.
  Springer, 2004.

\bibitem{seeger2003fast}
Matthias Seeger, Christopher Williams, and Neil Lawrence.
\newblock Fast forward selection to speed up sparse gaussian process
  regression.
\newblock In {\em Artificial Intelligence and Statistics 9}, number
  EPFL-CONF-161318, 2003.

\bibitem{titsias2009variational}
Michalis~K Titsias.
\newblock Variational learning of inducing variables in sparse gaussian
  processes.
\newblock In {\em AISTATS}, volume~12, pages 567--574, 2009.

\bibitem{lawrence2003fast}
Neil Lawrence, Matthias Seeger, and Ralf Herbrich.
\newblock Fast sparse gaussian process methods: The informative vector machine.
\newblock In {\em Proceedings of the 16th Annual Conference on Neural
  Information Processing Systems}, number EPFL-CONF-161319, pages 609--616,
  2003.

\bibitem{saatcci2012scalable}
Yunus Saat{\c{c}}i.
\newblock {\em Scalable inference for structured Gaussian process models}.
\newblock PhD thesis, University of Cambridge, 2012.

\bibitem{wilson2014fast}
Andrew Wilson, Elad Gilboa, John~P Cunningham, and Arye Nehorai.
\newblock Fast kernel learning for multidimensional pattern extrapolation.
\newblock In {\em Advances in Neural Information Processing Systems}, pages
  3626--3634, 2014.

\bibitem{wilson2015kernel}
Andrew Wilson and Hannes Nickisch.
\newblock Kernel interpolation for scalable structured gaussian processes
  (kiss-gp).
\newblock In {\em Proceedings of The 32nd International Conference on Machine
  Learning}, pages 1775--1784, 2015.

\bibitem{deisenroth2015gaussian}
Marc~Peter Deisenroth, Dieter Fox, and Carl~Edward Rasmussen.
\newblock Gaussian processes for data-efficient learning in robotics and
  control.
\newblock {\em IEEE Transactions on Pattern Analysis and Machine Intelligence},
  37(2):408--423, 2015.

\bibitem{rasmussen2006gaussian}
Carl~Edward Rasmussen and Christopher~KI Williams.
\newblock Gaussian processes for machine learning.
\newblock 2006.

\bibitem{snelson2005sparse}
Edward Snelson and Zoubin Ghahramani.
\newblock Sparse gaussian processes using pseudo-inputs.
\newblock In {\em Advances in neural information processing systems}, pages
  1257--1264, 2005.

\bibitem{wilson2015thoughts}
Andrew~Gordon Wilson, Christoph Dann, and Hannes Nickisch.
\newblock Thoughts on massively scalable gaussian processes.
\newblock {\em arXiv preprint arXiv:1511.01870}, 2015.

\end{thebibliography}

\newpage
\appendix
\section{Sparse Low-rank Kernel Approximation}
Algorithm \ref{alg:Sparse Low-rank Kernel Approximation} outlines our distillation approach.

\begin{algorithm}[h]
\caption{Sparse Low-rank Kernel Approximation}
\label{alg:Sparse Low-rank Kernel Approximation}
\begin{algorithmic}[1]
\State \textbf{Input:} A well trained kernel function $k_\gamma$, training feature vectors $X$ and targets $\bfy$, step size $\eta$, number of iterations $T$ and sparsity $b$. 
\State \textbf{Output:} Approximated kernel matrix $W K_{U,U} W^\top$
\State $U\gets \textsc{K-Means}(X)$
\State $\KXX \gets k_\gamma(X, X)$
\State $\KUU \gets k_\gamma(U, U)$
\State $\KXU \gets k_\gamma(X, U)$
\State \textit{Step 1: Initialization}
\State $W \gets \mathbf{0} \in \mathcal{R}^{n\times m}$
\For{each $\bfx_i$ in $X$}
\State $J\gets\text{indices for } b \text{ nearest neighbors of }\bfx_i\text{ in }U$
\State $W_{i}(J)\gets \text{argmin}_\beta ||\beta\KUU(J)- (\KXU)_i||_2$ 
\EndFor
\State \textit{Step 2: Gradient Descent}
\For{$t = 1$ to $T$}
\State $E\gets W\KUU W^\top - \KXX$
\State $E_{i,i} \gets 2E_{i,i} \text{ for } 1\le i\le n$
\State $\nabla W \gets$ $E^\top W\KUU$
\State Project each row of $\nabla W$ to $b$-sparse space
\State Update $W\gets W - \eta\nabla W$
\EndFor
\end{algorithmic}
\end{algorithm}

\section{Simulation Experiment}
\paragraph{Kernel Reconstruction.}
\begin{figure*}[h]
\centering
\begin{tabular}{cccc}
\includegraphics[width=0.23\textwidth]{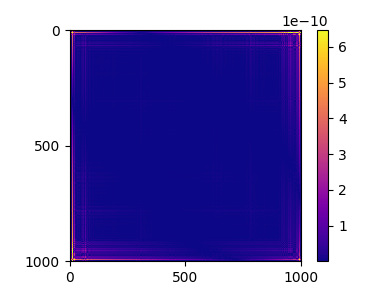} &
\includegraphics[width=0.23\textwidth]{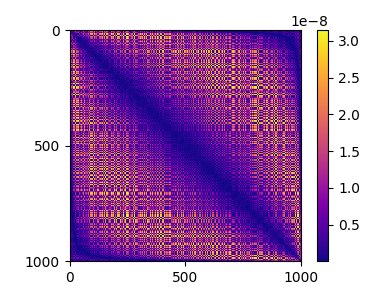} &
\includegraphics[width=0.23\textwidth]{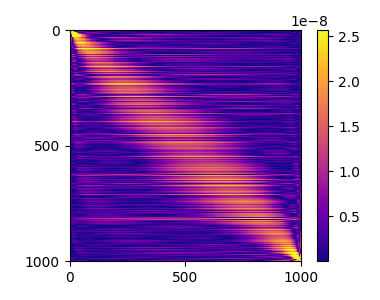} &
\includegraphics[width=0.22\textwidth]{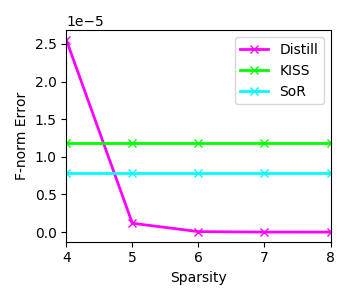}
\\
(a) $|\KXX - K_\text{distill}|$ & (b) $|\KXX- K_\text{KISS}|$ &
(c) $|\KXX - K_\text{SoR}|$ & (d) Error v.s. $b$
\end{tabular}
\caption{Kernel Reconstruction Experiments. (a) - (c) Absolute error matrix for reconstructing $\KXX$ with kernel distillation, KISS-GP and SoR respectively. (d) $F$-norm error for reconstructing $\KXX$ with distillation under different setting of $b$ (sparsity constraint) for $W$.}
\label{fig:kernel_reconstruction}
\end{figure*}

We first study how well can kernel distillation reconstruct the full teacher kernel matrix. We generate a 1000 $\times$ 1000 kernel matrix $\KXX$ from RBF kernel evaluated at (sorted) inputs $X$ randomly sampled from $\mathcal{N}(0,25)$. We compare kernel distillation against KISS-GP and SoR (FITC is essentially SoR with diagonal correction as mentioned in Section 2). We set number of grid points for KISS-GP as 400 and number of inducing points for SoR is set to 200 and kernel distillation to 100. We set the sparsity $b$ to 6 for kernel distillation. 

The $F$-norm for errors for are
$1.22\times10^{-5}$, $8.17\times 10^{-6}$, $2.39\times 10^{-7}$ for KISS-GP, SoR and kernel distillation respectively. Kernel distillation achieves lowest $F$-norm error compared to FITC and KISS-GP even the number of inducing points is much fewer for kernel distillation. Moreover, from the absolute error matrices (Figure~\ref{fig:kernel_reconstruction} a-c), we can see errors are more evenly distributed for kernel distillation, while there seems to exist a strong error pattern for the other two. 

We also show how the sparsity parameter $b$ affect the approximation quality. We evaluate the error with different choices for $b$ as shown in Figure~\ref{fig:kernel_reconstruction} (d). We observe that the error converges when the sparsity $b$ is above 5 in this example. This shows our structured student kernel can approximate the full teacher kernel reasonably well even when $W$ is extremely sparse.

\paragraph{Toy 1D Example.}
\begin{figure}[t]
\centering
\begin{tabular}{cc}
\includegraphics[width=0.22\textwidth]{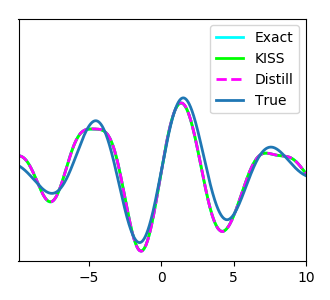} &
\includegraphics[width=0.22\textwidth]{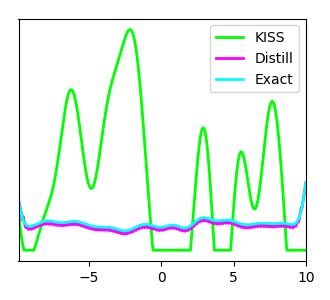} 
\\
(a) Mean & (b) Variance
\end{tabular}
\caption{Mean (a) and variance (b) prediction comparison for KISS-GP and Kernel Distillation on 1D example.}
\label{fig:1D_toy_example}
\end{figure}

To evaluate our distilled model's predictive ability, we set up the following experiment. We sample $n = 1000$ data points $X$ uniformly from [-10, 10]. We set our response $y(x) = \sin(x)\exp(-\frac{x^2}{2\times 5^2}) +\epsilon$ with $\epsilon \sim \mathcal{N}(0,1)$. We train an exact GP with RBF kernel as teacher first then apply kernel distillation with number of inducing points set to 100 and sparsity set to 10. We compare mean and variance prediction of kernel distillation with KISS-GP trained with 400 grid inducing points.

The results are showed in Figure~\ref{fig:1D_toy_example}. As we can see, mean predictions of kernel distillation are indistinguishable from exact GP and KISS-GP. As for variance, kernel distillation's predictions are much closer to the variance outputs from exact GP, while the variance outputs predicted by KISS-GP are far away from the exact solution.

This experiment shows a potential problem in KISS-GP, where it sacrifices its ability to provide uncertainty measurements,  which is a crucial property of Bayesian modeling, for exchanging massive scalability. On the other hand, kernel distillation can honestly provide uncertainty prediction close to the exact GP model.

\section{Experiment Setup} We compare kernel distillation with teacher kernel (exact GP), FITC as well as KISS-GP. We use the same inducing points selected by K-Means for both FITC and kernel distillation. For KISS-GP, as all the datasets do not lie in lower dimension, we project the input to 2D and construct 2D grid data as the inducing points. Number of inducing points (on 2D grid) for KISS-GP are set to 4,900 (70 per grid dimension) for Boston Housing, 10K for Abalone, 90K for PUMADYM32N, 250K for KIN40K. The number of inducing points for FITC and kernel distillation are 70 for Boston Housing, 200 for Abalone, 1k for PUMADYM32N and KINK40K. The sparsity $b$ in kernel distillation is set to 20 for Boston Housing and 30 for other datasets. For all methods, we choose ARD kernel as the kernel function, which is defined as:
\[k_\text{ARD}(\bfx, \mathbf{z}) = \exp(-0.5\sum_{i=1}^d (\bfx_i - \mathbf{z}_i)^2/\sigma_i^2)\]
where $d$ is the dimension of the input data and $\sigma_i$'s are the hyper-parameters to learn.

All the experiments were conducted on a PC laptop with Intel Core(TM) i7-6700HQ CPU @ 2.6GHZ and 16.0 GB RAM.

\end{document}